\documentclass[letterpaper, 10 pt, conference]{ieeeconf}

\IEEEoverridecommandlockouts  % RA-L 初投必须保留：开启 \thanks 命令
\ifdefined\pdfobjcompresslevel\pdfobjcompresslevel=0\fi
\usepackage{xcolor}

\usepackage{amsmath,amssymb,amsfonts}
\usepackage{graphicx}
\usepackage{booktabs}
\usepackage{url}
\usepackage{cite}
\usepackage{microtype}
\usepackage{tikz}

\usetikzlibrary{arrows.meta,calc,positioning}
\definecolor{annblue}{RGB}{47,110,171}
\tikzset{
    annline/.style = {draw=annblue, line width=0.4pt, line cap=round},
    annbox/.style  = {draw=annblue, line width=0.4pt, fill=white,
                      rounded corners=0.6pt, inner sep=1pt,
                      font=\tiny\bfseries, text=black},
    anndot/.style  = {circle, draw=annblue, line width=0.4pt, fill=annblue,
                      inner sep=0pt, minimum size=1.6pt},
}

\makeatletter
\@ifundefined{l@en}{\expandafter\def\csname l@en\endcsname{0}}{}
\makeatother

\newcommand\subfiglabel[1]{{\footnotesize (#1)}}

\newif\ifanonymous
\anonymousfalse

\title{\LARGE \bf
GAP: Generalized Adaptation Prior for Dexterous Piano Playing}% chktex 13：标题里「缩写: 大写词」属规则 13 误报

\ifanonymous
\author{\vspace{1.5\baselineskip}}
\else
\author{Jinyi Chen$^{1}$, Hongyi Li$^{1}$, Lei Jiang$^{1}$, Xianwei Liu$^{3}$, Yanhong Liang$^{1}$, \\ Yulin Chen$^{1}$, Yucheng Tao$^{1}$, Yongbin Jin$^{2,3\dagger}$, and Hongtao Wang$^{1,2\dagger}$%
\thanks{$^{\dagger}$Corresponding authors: Yongbin Jin and Hongtao Wang.}%
\thanks{$^{1}$Center for X-Mechanics, Zhejiang University, Hangzhou, China.}%
\thanks{$^{2}$ZJU-Hangzhou Global Scientific and Technology Innovation Center, Hangzhou, China.}%
\thanks{$^{3}$Mirrorme Technology Co., Ltd., Shanghai, China.}%
}
\fi

\makeatletter
\@ifundefined{IEEEkeywords}{%
        {\par\vspace{4pt}\noindent\textit{Index Terms}---\relax}%
        {\par}%
}{}
\makeatother

\begin{document}

% 关闭 IEEEtran.bst 对 url 字段的输出，避免每条文献末尾多出一行
%   "[Online]. Available: \url{https://...}"（共 31 条，明显占版面）。
% 控制条目定义在 references.bib 末尾，详见那里的注释。
% 想恢复在线链接：注释掉下面这一行，再重跑 bibtex 即可。
\bstctlcite{IEEEtranBSTControl}

\maketitle
\thispagestyle{empty}
\pagestyle{empty}

% 本图原先在此处以 \begin{figure*}[t] 声明，实际上会被顺延到第 2 页顶部。
% 现已改为首页 teaser：图与图题定义在导言区的 \FappTeaser 宏中，
% 由 \IEEEaftertitletext 注入标题区，展开后位于标题/作者与摘要正文之间，
% 图题 \label 沿用 fig:fig_real_adaptation。

%%%%%%%%%%%%%%%%%%%%%%%%%%%%%%%%%%%%%%%%%%%%%%%%%%%%%%%%%%%%%%%%%%%%%%%%%%%%%%%%
% 陈述问题、你的方法与解、论文的主要贡献；背景和动机尽可能少；要实事求是但全面，并且摘要里的材料不应在后文逐字重复
% 单段、≤250 词、自含、无缩写/脚注/文献/公式、点出新意、附 3–5 个关键词，且不要在标题和摘要里放数学符号，关键词要尽量用标准化词，可以用 IEEE Thesaurus 去找
% 要避免的是把摘要写成“方法目录”：那里不该出现“首先…然后…最后…”。
\begin{abstract}
    % 摘要逻辑：灵巧操作应用，问题，点出让人印象深刻的中心贡献，方法，结果，展望
    Dexterous hands enable versatile manipulation, but policies trained in simulation often degrade on hardware owing to non-idealities, and the high-dimensional action space makes real-world adaptation costly. % 应用-问题
    Piano playing, which requires precise and coordinated multifinger motion, offers a representative testbed.
    % 适当减少方法篇幅
    We propose \textbf{GAP} (\underline{G}eneralized \underline{A}daptation \underline{P}rior for Dexterous Piano Playing), a sample-efficient adaptation framework
    that learns how to adapt, rather than how to act, during simulation pretraining and transfers this adaptation prior to hardware.
    Between repeated executions, GAP uses feedback to adjust the actions of the next one:
    a gradient guidance network learns from paired executions how adjusting actions affects playing performance, and its gradients guide these adjustments.
    % \yanhongComment{At least using one sentence to point out the main method.}
    In simulation, GAP yields approximately 2.4 times the F1 scores improvement of the existing fast-adaptation baseline for dexterous piano playing under the same interaction budget. The learned prior also generalizes to musical pieces held out from pretraining.
    On a physical dexterous hand, the same simulation-pretrained prior improves F1 scores by 3.6\%--23\% across three pieces of varying difficulty after approximately 7 min of adaptation each.
    These results suggest a practical route to reduce the real-world interaction needed to deploy dexterous hands.
    % 重点给结果，要量化，讲3个点，与引言的3个贡献呼应
    % 摘要8句话，3个结果

\end{abstract}

\section{Introduction}
\label{sec:intro}

% 逻辑链：为什么要研究灵巧操作？(应用场景) -> 硬件能力已具备 -> 瓶颈在实机适应 -> 钢琴暴露该瓶颈且真机交互昂贵 -> 应该从仿真寻求交互 -> 仿真该迁移什么 -> 本文问题
Dexterous hands are expected to bring human-like versatility to robots, from tool use, assembly to handling deformable objects and playing musical instruments.
The hardware has advanced rapidly, with greater force output, faster motion, and higher precision\cite{shaw_leap_2023,zhang_biomimetic_2025}.
Turning this capability into reliable behavior, however, depends on how well a skill matches the physical hand\cite{zhao_dexctrl_2026}.
Friction, hysteresis, transmission compliance, and actuator delay are captured only approximately by nominal models and simulators, and the resulting deviations degrade real-world performance\cite{yuan_development_2026,gao_simtoreal_2024,aljalbout_reality_2026}.
The practical value of dexterous hands therefore depends increasingly on how efficiently their skills can be adapted to the physical system.

Piano playing exposes this need clearly\cite{zakka_robopianist_2023}.
Every keystroke must reach the correct key, with sufficient depth, at the right time, while several fingers move in coordination; deviations of a few millimeters can produce wrong or missed notes\cite{xie_handelbot_2026,zeulner_learning_2026}.
Systems that rely on accurate calibration and modeling can play physical pianos, but they cannot account for deviations that remain unmodeled.
Real-world feedback reveals these errors, but correcting them is difficult: the actions are high-dimensional, the outcome consists of discrete keystroke events whose dependence on the actions is hard to model, and every trial occupies the hardware.
Residual reinforcement learning (RL)\cite{johannink_residual_2019,xie_handelbot_2026} learns such corrections from real-world returns, which still requires extensive real-world interaction\cite{tang_deep_2025a,levy_simulation_2026}.

Simulation is the natural source of inexpensive experience, but the question is what to transfer.
Policies and dynamics models learned in simulation are fitted to the simulated dynamics, which is precisely what differs on hardware\cite{levy_simulation_2026,hsieh_learning_2026}.
What may remain valid despite this mismatch is how an adjustment changes performance relative to an observed execution: when a fingertip lands off center, shifting it toward the key center improves the keystroke, whichever piece is played and whatever caused the offset\cite{xie_handelbot_2026,jiang_transic_2025}.
This raises the question addressed in this letter: can such a \emph{generalized adaptation prior}, learned from diverse simulated executions, substantially reduce the real-world interaction required for adaptation?

% 审稿人七问当写作清单：论文的贡献是什么？作者是否阐明了该论文的重要性？论文写作是否清晰、结构是否合理？引言是否明确了论文的研究目的？参考文献是否相关且完整？如有缺失，请补充。如果论文在技术层面不够严谨，原因何在？如果论文篇幅过长，应如何精简？

% 点出论文的核心贡献，找到一种方法，能够把这个通用适应先验在仿真中学到，并迁移到现实以减少真机交互需求
% 重要性：能够利用仿真中廉价的交互

\begin{figure}
    \centering
    \includegraphics[width=\columnwidth]{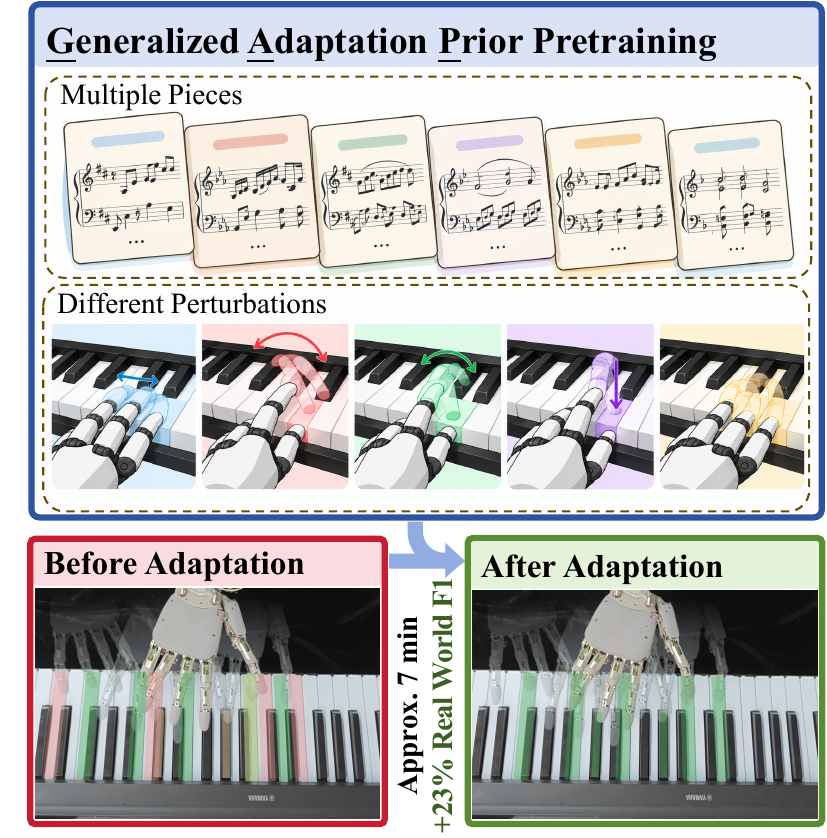}
    \vspace{-22pt}
    \caption{GAP pretrains a generalized adaptation prior in simulation on multiple pieces under different perturbations,
        including joint zero offsets, damping, actuator delays, and sensor noise (top),
        and uses it for fast adaptation on the physical hand (bottom).
        Green, red, and orange keys denote correct, incorrect, and missed keystrokes.}
    \label{fig:first_page_figure}
    \vspace{-15pt}
\end{figure}

Building on this idea, we propose GAP (Generalized Adaptation Prior), a sample-efficient adaptation framework for dexterous piano playing (Fig.~\ref{fig:first_page_figure}).
Rather than learning how to act, GAP learns how to adapt: a Gradient Guidance Network (GGN) scores a correction to a given initial trajectory by how it would change keystroke accuracy within a short time window relative to an observed execution of that trajectory.
GGN is pretrained on simulated executions of multiple pieces under diverse perturbations and then refined on hardware with executions of each piece.
In simulation, GAP improves the F1 score approximately 2.4 times as much as the existing fast-adaptation method for dexterous piano playing under the same interaction budget, pretraining cuts the needed interaction by about $4.4$ times, and the prior generalizes to unseen pieces.
On hardware, the same prior raises F1 by $3.6\%$--$23\%$ in about $7$\,min per piece, and continual fine-tuning yields further gains.

The main contributions of this work are as follows:

\begin{itemize}
    % 贡献固定3点，句式一致，且和摘要的3个结果呼应
    % 第一条讲框架贡献，能够利用仿真中廉价的交互，把这个通用适应先验在仿真中学到，并迁移到现实以减少真机交互需求,与论文的核心洞察对应
    % 第二条讲具体实现的贡献，GGN 能够把离散的琴键事件反馈，变成引导动作修正的梯度，与钢琴适应的难点对应
    % 第三条讲实验验证，在仿真和真机实验中，证明了GAP的样本高效和快速适应，在不同曲目和偏差下的泛化性
    \item We propose \textbf{GAP}, a framework that learns a generalized adaptation prior from inexpensive simulated executions and transfers it to hardware to reduce real-world interaction.
    \item We develop the GGN, which turns discrete keystroke feedback from a completed execution into gradients that adjust the actions of the next one.
    \item We demonstrate in simulation and real-world experiments that GAP adapts sample-efficiently and generalizes across musical pieces and execution deviations.
\end{itemize}

%%%%%%%%%%%%%%%%%%%%%%%%%%%%%%%%%%%%%%%%%%%%%%%%%%%%%%%%%%%%%%%%%%%%%%%%%%%%%%%%
%%%%%%%%%%%%%%%%%%%%%%%%%%%%%%%%%%%%%%%%%%%%%%%%%%%%%%%%%%%%%%%%%%%%%%%%%%%%%%%%
%%%%%%%%%%%%%%%%%%%%%%%%%%%%%%%%%%%%%%%%%%%%%%%%%%%%%%%%%%%%%%%%%%%%%%%%%%%%%%%%
%%%%%%%%%%%%%%%%%%%%%%%%%%%%%%%%%%%%%%%%%%%%%%%%%%%%%%%%%%%%%%%%%%%%%%%%%%%%%%%%
%%%%%%%%%%%%%%%%%%%%%%%%%%%%%%%%%%%%%%%%%%%%%%%%%%%%%%%%%%%%%%%%%%%%%%%%%%%%%%%%
\section{Related Work}
\label{sec:related}

% 每个子小节的写法（四步）
% 分类陈述：先给该类方法的共同范式或代表工作。
% 归纳进展：用一句话概括这一族做到了什么。
% 指出局限。
% 定位本文：说明本文如何不同/更进一步（也可放在小节末或 Related Work 结尾统一收口）。

% 强调：近 2–3 年的期刊论文（官方评审表明确建议）；与本文最接近的 2–3 篇必须逐一对比， 说明差异；引用完整、无遗漏。
% 避免：只罗列“A 做了… B 做了…”的流水账；只引会议和 arXiv；把 Related Work 写成背景科普 而没有批判。
% 务实：RA-L 版面紧张，宁可少而准——按 RAL参考文献规律分析.md 的结论，全文参考文献 30–40 篇较稳妥。

% 相关工作这节应该突出论文的创新点，同时不要让人误解为论文的方法是RL或sim2real

\subsection{Dexterous Piano Playing}

Research on dexterous piano playing has progressed from simulation benchmarks to real-world deployment and adaptation.
RoboPianist\cite{zakka_robopianist_2023} demonstrated piano playing through RL in simulation and introduced framewise precision, recall, and F1 metrics, which we adopt in this work.
Wang et al.\cite{wang_human_2024}, Liang et al.\cite{liang_expressive_2026} and Jiang et al.\cite{jiang_biomimetic_2026} demonstrated physical piano playing, but their reliance on accurate calibration and modeling limits adaptation for unmodeled real-world deviations.
Zeulner et al.\cite{zeulner_learning_2026} use real-world data to update simulator parameters in a Sim2Real2Sim loop, but each round retrains the policy in simulation and redeploys it, which does not provide fast adaptation to a given piece.
HandelBot\cite{xie_handelbot_2026} incorporates real-world data into the adaptation loop through heuristic lateral-joint correction of open-loop trajectories followed by real-world residual RL\@.
In contrast, GAP does not use residual RL and addresses these deployment deviations without manually designed lateral-alignment rules. % 与 residual RL 的对比在下面

\subsection{Fast Real-World Adaptation}

Residual RL reduces the difficulty of real-world learning by learning residual actions on top of a base controller\cite{johannink_residual_2019, li_hybrid_2025a, lin_multimodal_2025}.
However, rollout costs, observation noise, and temporal credit assignment in multicontact, long-horizon, and high-degree-of-freedom tasks
still necessitate extensive real-world interaction and limit sample efficiency\cite{zhang_efficient_2023, elguea-aguinaco_review_2023, tang_deep_2025a}.
Rather than inferring the effect of a correction from long-horizon returns, GAP measures the local reward change induced by a correction from paired executions that differ only in that correction, and trains GGN on the measurements by supervised learning instead of trial-and-error credit assignment.

Iterative learning control (ILC) updates feedforward inputs based on previous executions,
providing a framework for adapting repetitive tasks without an accurate model\cite{arimoto_bettering_1984,zhang_neuralnetworkbased_2021}.
For example, Chi et al.\cite{chi_iterative_2024} learn incremental dynamics relating action increments to trajectory changes and use them to optimize dynamic manipulation of deformable objects.
GAP similarly learns from how action corrections affect a previous execution, but it learns reward changes computed from discrete keystroke events rather than trajectory changes, and updates action corrections through gradient guidance rather than action sampling.

\subsection{Simulation Priors for Real-World Transfer}

Simulation priors differ in what they transfer to hardware.
Policies trained with domain randomization can be deployed directly\cite{bauza_demostart_2025,kedia_simtoolreal_2026} or used to initialize real-world RL\cite{wu_simlauncher_2025,hu_slac_2025}.
Dynamics and value models learned in simulation guide planning, policy updates, or exploration on hardware\cite{lanier_redraw_2025,levy_simulation_2026,yin_rapidly_2025}.
GAP instead transfers a relative prior: GGN learns a surrogate that scores action corrections by their reward change relative to an observed execution of the initial trajectory.
On hardware, this execution is a real one, so deviations of the physical hand enter through the observations rather than through the prior, and the same prior is shared across pieces.

\begin{figure*}[t]
    \centering
    \includegraphics[width=\textwidth]{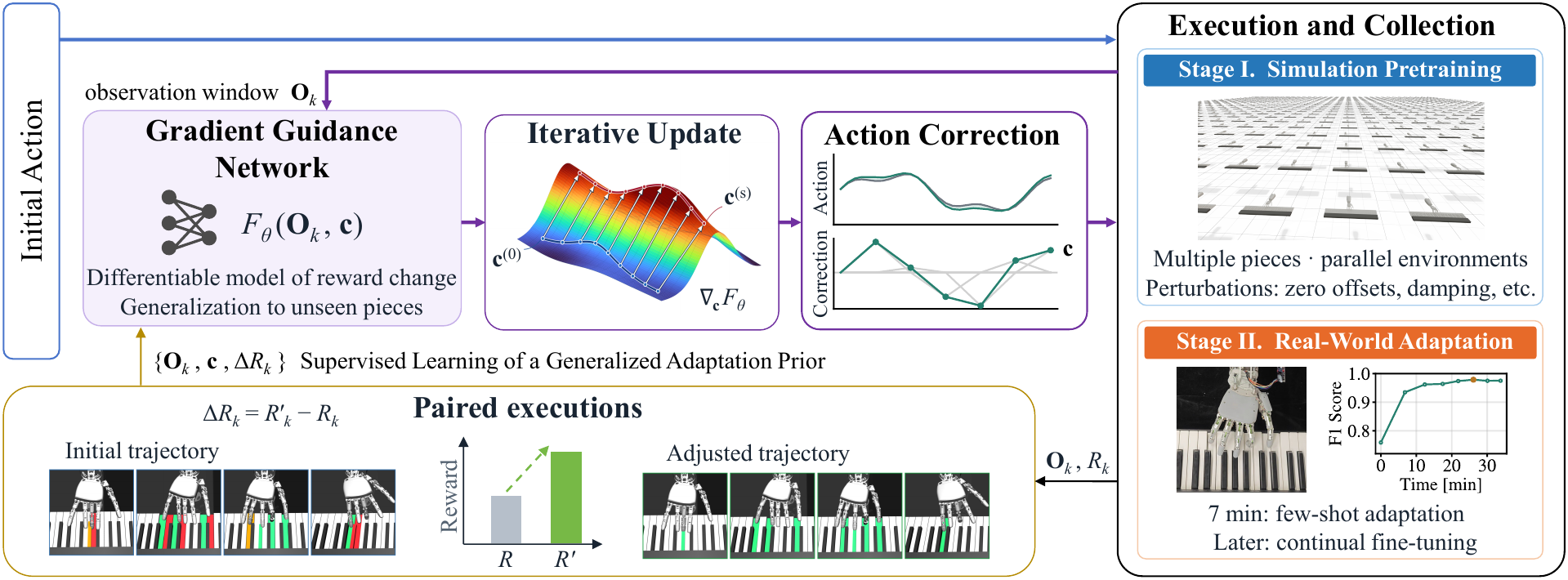}
    \vspace{-20pt}
    \caption{\textbf{Overview of GAP\@.}
        Both stages share one adaptation process: gradients of the reward change modeled by GGN $F_\theta$ iteratively update the action corrections, and paired executions of the initial and adjusted trajectories supervise GGN\@.
        The execution and collection module is instantiated in parallel simulation in Stage I and on the physical hand in Stage II\@.}
    \label{fig:method_overview}
    \vspace{-15pt}
\end{figure*}

%%%%%%%%%%%%%%%%%%%%%%%%%%%%%%%%%%%%%%%%%%%%%%%%%%%%%%%%%%%%%%%%%%%%%%%%%%%%%%%%
%%%%%%%%%%%%%%%%%%%%%%%%%%%%%%%%%%%%%%%%%%%%%%%%%%%%%%%%%%%%%%%%%%%%%%%%%%%%%%%%
%%%%%%%%%%%%%%%%%%%%%%%%%%%%%%%%%%%%%%%%%%%%%%%%%%%%%%%%%%%%%%%%%%%%%%%%%%%%%%%%
%%%%%%%%%%%%%%%%%%%%%%%%%%%%%%%%%%%%%%%%%%%%%%%%%%%%%%%%%%%%%%%%%%%%%%%%%%%%%%%%
%%%%%%%%%%%%%%%%%%%%%%%%%%%%%%%%%%%%%%%%%%%%%%%%%%%%%%%%%%%%%%%%%%%%%%%%%%%%%%%%
\section{Method}
\label{sec:method}

% 行文结构
% Overview（1–2 段 + 一张 pipeline 图）：先用文字和系统图讲清“输入→处理→输出”，让读者在 看公式前建立全局观。范文几乎都有 Fig. 1/2 的 pipeline 或系统总览。
% Problem Formulation / Preliminaries（可选单设小节）：定义符号、状态、目标函数，写清假设 与边界条件。控制/建模类稿件常把 Simulation Preliminaries 单列。
% 分小节展开模块：每个模块对应贡献列表里的一条；先动机（为什么这样设计），再形式化（公式）， 最后实现细节（网络结构、超参、训练流程）。
% Training / Implementation Details：训练数据规模、损失、训练方法、奖励设计、域随机化设置等， 保证可复现。
% 要强调 / 要避免
% 强调：每个设计选择都有理由（对应某个缺口）；符号一致；公式必要且编号；给出可复现所需的 关键参数。官方对 Methodology 的定义就是“straightforward + reproducible”。
% 避免：公式堆砌而无动机；把实现细节写成流水账；方法节引用过多文献（范文方法节通常只引 2–6 篇算法来源/背景）。
% 长度参考：方法/建模是全文最长章节之一；控制类可配 15–30 个编号公式，硬件设计类可接近 0。

As shown in Fig.~\ref{fig:method_overview}, GAP has two stages: Stage I learns the prior from diverse simulated executions (Section~\ref{sec:pretrain}), and Stage II applies and refines it on hardware (Section~\ref{sec:realworld}).
Both stages share the same adaptation process, which takes place between repeated executions rather than within one.
GGN takes observations from a completed execution and action corrections as inputs, and outputs a scalar score: the local reward change these corrections would induce relative to that execution.
With the observations held fixed, the corrections are iteratively updated along the gradients of this score with respect to them.
Executing the adjusted trajectory yields paired feedback for network training.
Stage I outputs only the GGN parameters, which constitute the adaptation prior; Stage II adapts each piece independently, starting from this prior.

\subsection{Gradient Guidance Network}
\label{sec:ggn}

% 这里要写点什么？还是空着？

\textbf{Observation.} Based on the sensors available on hardware, the observation at time step $t$ is $\mathbf{o}_t=\{\mathbf{a}_t, \mathbf{q}_t, \dot{\mathbf{q}}_t, \mathbf{y}_t, \mathbf{y}^{\star}_t, \mathbf{f}_t \}$,
where $\mathbf{a}_t \in \mathbb{R}^{d_a}$ is the target joint position command, $\mathbf{q}_t,\dot{\mathbf{q}}_t \in \mathbb{R}^{d_a}$ are the joint positions and velocities; $d_a$ is the action dimension;
$\mathbf{y}_t,\mathbf{y}^{\star}_t\in\{0,1\}^{88}$ indicate the activated and target keys of the 88-key piano at the time step,
and $\mathbf{f}_t \in \{0,1\}^{d_f}$ marks the fingers assigned to the target keys by a predefined fingering ($d_f=5$ for one hand and $10$ for two).

\textbf{Action Correction.} We adapt the initial trajectory by adding \emph{low-frequency action corrections}:
the initial trajectory $\mathbf{a}_{0:N-1} = \{\mathbf{a}_0,\mathbf{a}_1,\dots,\mathbf{a}_{N-1}\}$, where $N$ is the trajectory length,
is sent at $50\,\mathrm{Hz}$ and tracked by joint-level control loops,
whereas action corrections are specified at $10\,\mathrm{Hz}$ on correction nodes.
The choice of correction frequency is supported by the experiments in Section~\ref{sec:exp_design}.
The action correction $\Delta \mathbf{a}_{t}$ at each time step is obtained by interpolating the correction nodes:

\begin{equation}
    \Delta \mathbf{a}_{t} = \sum_{k=0}^{M-1} \mathbf{c}_{k}
    \cdot \underbrace{\max\left(0,\; 1 - \frac{|t-t_k|}{T_{c}}\right)}_{\text{interpolation weight}}
    \label{eq:interp}
\end{equation}

where $\mathbf{c}_k \in \mathbb{R}^{d_a}$ is the correction at node $k$ (time $t_k$) and $M$ is the number of nodes; $T_c$ denotes the number of time steps between correction nodes, with $T_c = 5$ in this work; the adjusted trajectory is $\mathbf{a}'_{t} = \mathbf{a}_{t} + \Delta \mathbf{a}_{t}$.

\textbf{Local Reward Change.}
Following RoboPianist\cite{zakka_robopianist_2023}, we quantify playing performance using precision, recall, and their harmonic mean, F1, computed from $\mathbf{y}_t$ and $\mathbf{y}^{\star}_t$ .
Precision is the fraction of activated keys that are correct, and recall is the fraction of target keys that are activated; each is set to $1$ when its denominator is zero.
The whole-piece F1 score is the mean of the framewise $\mathrm{F1}_t$ scores.
The reward at each time step $t$ is $r_{t} = (\mathrm{Precision}_t + \mathrm{Recall}_t)/2$.
In a \emph{paired execution}, the initial and adjusted trajectories are executed sequentially in the same environment (with the same perturbation configuration in simulation), giving $\Delta r_{t} = r'_{t} - r_{t}$, where $r'_{t}$ and $r_{t}$ are their respective rewards.
The local reward change at correction node $k$ is

% 这里没有解释为什么不把F1作为奖励。
% 是因为之前做过对比实验，当前奖励形式效果好于直接用F1，但无法有信服的解释，
% 论文中也没有该对比实验用于证明，干脆不提

\begin{equation}
    \Delta R_{k} = \sum_{t=t_k-T_c}^{t_k+T_c} \Delta r_t \cdot
    \underbrace{\left[\frac{1}{T_c}
            \left(1 - \frac{|t-t_k|}{T_c}\right)\right]}_{\text{normalized weight}}
    \label{eq:localreward}
\end{equation}
The weights in~\eqref{eq:localreward} match the triangular interpolation kernel in~\eqref{eq:interp}, temporally aligning each correction node with its evaluation window.
Using differences rather than absolute scores focuses GGN on the \emph{change} induced by an action correction relative to an existing execution and gives samples from all pieces and perturbations a common supervision target, without modeling the full dynamics.

\textbf{Network Formulation and Training.}
For network training and gradient-guided adaptation, we use dimensionless correction variables $\tilde{\mathbf{c}}_k$, related to the physical corrections by $\mathbf{c}_k=\mathbf{D}\tilde{\mathbf{c}}_k$, where the diagonal $\mathbf{D}$ scales wrist, finger, and arm translational joints by $0.02$ rad, $0.1$ rad, and $0.002$ m, respectively.
The network input at node $k$ includes a local observation window

\vspace{-10pt}
\begin{equation}
    \mathbf{O}_{k} = \mathbf{o}_{t_k-f_o:t_k+f_o}
    \label{eq:localobs}
\end{equation}
% 大写 O 表示窗口，避免与逐帧观测 o_t 混用

with radius $f_o = 10$. Because the window is taken from a completed execution, it can include steps after $t_k$; this bidirectional window outperforms a unidirectional one (Section~\ref{sec:exp_design}).
GGN is formulated as $\widehat{\Delta R}_k = F_\theta(\mathbf{O}_k, \tilde{\mathbf{c}}_{k-n_c:k+n_c})$,
where $F_\theta$ is a neural network parameterized by $\theta$ with scalar output $\widehat{\Delta R}_k$.
The inputs to $F_\theta$ include not only the local observation $\mathbf{O}_k$ and dimensionless correction $\tilde{\mathbf{c}}_k$ at the current correction node,
but also the corrections at $n_c$ neighboring nodes on either side, $\tilde{\mathbf{c}}_{k-n_c:k+n_c}$:
neighboring nodes affect the actual action at the current time through the interpolation kernel in~\eqref{eq:interp}, and ignoring this coupling would introduce systematic bias.
We set $n_c=2$.
Paired executions yield the samples
$\mathcal{D} = \left\{(\mathbf{O}_k, \tilde{\mathbf{c}}_{k-n_c:k+n_c}, \Delta R_k) \;\middle|\; k = 0,1,\dots,M-1\right\}$,
which are used for supervised training:

\begin{equation}
    \min_\theta\; \mathbb{E}_{\mathcal{D}}
    \left[\left(F_\theta(\mathbf{O}_k,\tilde{\mathbf{c}}_{k-n_c:k+n_c})
        - \Delta R_k\right)^2\right]
    \label{eq:qtrain}
\end{equation}

% 常规超参可按篇幅取舍

\textbf{Architecture.} GGN $F_\theta$ is a $2$-layer pre-norm Transformer\cite{vaswani_attention_2017}
($128$-dimensional tokens, $4$ heads, feedforward dimension $256$).
Each time step of $\mathbf{O}_k$ and each correction node in $\tilde{\mathbf{c}}_{k-n_c:k+n_c}$ is embedded as a separate token with type and positional encodings;
attention pooling and a multilayer perceptron produce the scalar output, bounded to $[-0.5,0.5]$ by a scaled $\tanh$.

\textbf{Gradient-Guided Adaptation.} With execution observations and network parameters fixed, the corrections are updated along the gradients of the surrogate objective

\vspace{-10pt}

\begin{equation}
    J(\tilde{\mathbf{C}}) = \frac{1}{M}\sum_{k=0}^{M-1} \Bigl[
        \min_{j} F_{\theta_j}(\mathbf{O}_k, \tilde{\mathbf{c}}_{k-n_c:k+n_c})
        - \frac{\lambda}{d_a} \|\tilde{\mathbf{c}}_k\|_2^2 \Bigr]
    \label{eq:get_corr_action}
\end{equation}

Here, $\tilde{\mathbf{C}}=\tilde{\mathbf{c}}_{0:M-1}$ denotes the complete sequence of dimensionless correction nodes. The coefficient $\lambda$ penalizes correction magnitude, with $\lambda=0.01$ in Stage I and $0.02$ in Stage II\@.
We maintain two networks $F_{\theta_1},F_{\theta_2}$ with identical architectures but different initializations; each is trained independently with~\eqref{eq:qtrain} on the same paired samples. Taking the minimum network output via $\min_j$ is intended to reduce overestimation~\cite{fujimoto_addressing_2018}.
The gradients guide corrections toward improved execution rather than identifying an optimal trajectory under the true dynamics.

% 一个可能的质疑就是和轨迹优化的关系，解释：核心先验不是某一条固定的最优轨迹，而是执行观测、动作修正与表现变化之间的关系。

To bound the corrections, we parameterize $\tilde{\mathbf{C}}=\tanh(\mathbf{U})$ elementwise and update $\mathbf{U}$.
For random initialization, each component of $\mathbf{U}^{(0)}$ is drawn independently from $\mathcal{N}(0,0.1^2)$, or, in Stage II, may start from previous corrections (Section~\ref{sec:realworld}).
We then run $S=20$ Adam steps (learning rate $0.05$), updating $\mathbf{U}^{(s)}$ to $\mathbf{U}^{(s+1)}$ along $\mathbf{g}^{(s)} = \nabla_{\mathbf{U}} J(\tanh\mathbf{U})|_{\mathbf{U}^{(s)}}$, where gradients from overlapping local terms accumulate at each node.
The resulting $\tilde{\mathbf{C}}^{(S)}$ is scaled by $\mathbf{D}$, interpolated by~\eqref{eq:interp}, and added to the initial trajectory, and the adjusted trajectory is executed.

\subsection{Initial Trajectory}
\label{sec:train_init_traj}

We train a playing policy for each piece using the same RL procedure as Jiang et al.\cite{jiang_biomimetic_2026}.
The target position command sequence exported after training serves as the initial trajectory $\mathbf{a}_{0:N-1}$.
The playing policies are trained separately for each piece, whereas GGN is \emph{jointly pretrained} on adaptation experience from multiple pieces.

\begin{table}[t]
    \caption{Simulation perturbations}
    \centering
    \label{tab:perturbation}
    \footnotesize
    \vspace{-8pt}
    \begin{tabular}{lcc}
        \toprule
        Perturbation              & Hand (18 joints)                 & Arm (3 joints) \\
        \midrule
        \multicolumn{3}{l}{Per paired execution}                                      \\
        \quad Zero offset         & $\pm0.08$ rad$^{\dagger}$        & ---            \\
        \quad Damping scale       & \multicolumn{2}{c}{$[0.8,1.25]$}                  \\
        \quad Actuator delay      & $[2.5,30]$ ms                    & $[5,50]$ ms    \\
        \quad Action perturbation & $\pm0.005$ rad                   & $\pm0.005$ m   \\
        \midrule
        \multicolumn{3}{l}{Per time step}                                             \\
        \quad Position delay      & \multicolumn{2}{c}{$[0,20]$ ms}                   \\
        \quad Velocity delay      & \multicolumn{2}{c}{$[20,40]$ ms}                  \\
        \quad Position noise      & $\pm0.005$ rad                   & $\pm0.0002$ m  \\
        \quad Velocity noise      & $\pm1.0$ rad/s                   & $\pm0.02$ m/s  \\
        \bottomrule
    \end{tabular}

    \vspace{1pt}
    {\footnotesize $^{\dagger}$Wrist-yaw, thumb-base and abduction joints only. \\
        All ranges are uniform; $\pm a$ denotes $[-a,a]$.
        Action perturbations use $5$\,Hz interpolation nodes;
        the sampled sequence is fixed within each execution.}
    \vspace{-15pt}
\end{table}

\subsection{Stage I: Simulation Pretraining}% chktex 13
\label{sec:pretrain}

Stage I instantiates the execution and collection module in parallel simulation, training GGN $F_\theta$ across pieces and parameter perturbations to learn a generalized adaptation prior.
We divide the initial trajectory of each piece into $4$\,s segments and combine them into a shared sampling set. In each iteration, every parallel environment samples a segment
and resamples joint zero offsets, damping, and actuator delays to produce diverse execution deviations.
To improve GGN robustness, we additionally apply low-frequency perturbations to the initial action trajectories and add noise to sensor observations. Perturbation ranges are listed in Table~\ref{tab:perturbation}.
Pretraining alternates between gradient-guided adaptation and supervised updates. Each sampled initial trajectory is first executed to obtain the observation windows used by GGN-guided gradient ascent in~\eqref{eq:get_corr_action}.
For exploration, noise drawn from $\mathcal{N}(0,0.1^2)$ is added to the dimensionless corrections $\tilde{\mathbf{C}}^{(S)}$ produced by gradient ascent, and the noisy corrections are clipped to $[-1,1]$, the range of $\tanh$; the resulting adjusted trajectory is executed under the same perturbation configuration.
The paired results from the most recent $40$ iterations are kept in a sample buffer, from which training samples are drawn according to~\eqref{eq:qtrain}.
In both stages, correction nodes within $0.5$\,s of either segment end are excluded from training and from the network term in~\eqref{eq:get_corr_action}.
During the first iteration, before any paired data exist, random dimensionless corrections drawn from $\mathcal{N}(0,0.25^2)$ are used for cold-start exploration.
Parallel simulation supplies tens of thousands of segments for learning generalized adaptation guidance without the corresponding physical interactions.

\subsection{Stage II: Real-World Adaptation}% chktex 13：同上，规则 13 误报
\label{sec:realworld}

In Stage II, adaptation is performed on the physical robot, with each piece starting independently from the same simulation-pretrained GGN parameters. GGN gradients guide action corrections, while paired real-world executions of the current piece provide samples for fine-tuning to compensate for deviations insufficiently covered by simulation.
To control interaction costs, the full piece is divided into $4$\,s segments with $2$\,s overlap between adjacent segments. Executions and paired samples are formed per segment, whereas GGN is updated on samples pooled from all segments.
Before each segment execution, the hand moves to the command preceding the segment, with the arm lifted by $10$\,mm during the approach to avoid key contact.
In each iteration, segments that have not met the performance thresholds are visited in random order, with at most $3$ attempts per segment; a segment is skipped once F1 $>0.97$ and precision $>0.99$.
For each attempt, a cached execution of the initial trajectory is reused with probability $0.65$ (each cache entry is reused at most $3$ times); otherwise, the trajectory is executed again and observations are recorded.
Thus, throughout Stage II, the first execution of every pair is the same initial trajectory $\mathbf{a}_{0:N-1}$, and the observation windows $\mathbf{O}_k$ supplied to GGN always come from executions of this trajectory; corrections are therefore always defined relative to the initial trajectory rather than accumulated across iterations.
GGN-guided gradient ascent then produces action corrections using~\eqref{eq:get_corr_action}, and the adjusted trajectory is executed to form paired samples.
Action corrections are initialized randomly with probability $0.2$ and otherwise from one of the segment's previous corrections, sampled with softmax probabilities over the F1 scores those corrections achieved; in the first iteration, no previous corrections exist, so all initializations are random.
At the end of each iteration, GGN is trained for $12$ passes on samples from the most recent $4$ iterations, with at most $64$ minibatches of $1024$ correction-node samples per pass.
The current best corrections for individual segments are blended using linear weights in overlapping regions, and the full piece is executed to evaluate performance after the iteration.
All first-iteration action corrections use the pretrained GGN without real-world fine-tuning; the subsequent whole-piece evaluation defines the \textbf{few-shot} adaptation result.
Later iterations use GGN fine-tuned on the current piece's data for continual adaptation.

%%%%%%%%%%%%%%%%%%%%%%%%%%%%%%%%%%%%%%%%%%%%%%%%%%%%%%%%%%%%%%%%%%%%%%%%%%%%%%%%
%%%%%%%%%%%%%%%%%%%%%%%%%%%%%%%%%%%%%%%%%%%%%%%%%%%%%%%%%%%%%%%%%%%%%%%%%%%%%%%%
%%%%%%%%%%%%%%%%%%%%%%%%%%%%%%%%%%%%%%%%%%%%%%%%%%%%%%%%%%%%%%%%%%%%%%%%%%%%%%%%
%%%%%%%%%%%%%%%%%%%%%%%%%%%%%%%%%%%%%%%%%%%%%%%%%%%%%%%%%%%%%%%%%%%%%%%%%%%%%%%%
%%%%%%%%%%%%%%%%%%%%%%%%%%%%%%%%%%%%%%%%%%%%%%%%%%%%%%%%%%%%%%%%%%%%%%%%%%%%%%%%
\section{Experiments}
\label{sec:exp}

The experiments address three questions:
(i) which GGN design choices support adaptation (Section~\ref{sec:exp_design});
(ii) whether the simulation prior improves adaptation efficiency and generalizes to unseen pieces,
compared with the existing fast-adaptation method for dexterous piano playing and ablated variants (Section~\ref{sec:exp_sim}); and
(iii) whether the same prior transfers to the physical hand and supports continual adaptation
(Section~\ref{sec:exp_real}).

\subsection{Experimental Setup}
\label{sec:exp_setup}

% 核查 Beyond Hand 是否违反双盲
Both simulation and hardware experiments use a right-hand platform based on a tendon-driven dexterous hand\cite{jiang_data-driven_2026},
which has $2$ wrist joints and $16$ actuated finger joints and is mounted on a UR5 arm that provides $3$ translational degrees of freedom.
Actual keystrokes are recorded through MIDI messages from a YAMAHA P48 digital piano.
We evaluate on right-hand excerpts of approximately $30$\,s from Canon, Für Elise, and Beethoven's Symphony No.~5,
covering broken chords, stepwise motion, chromatic alternation, leaps, and repeated keystrokes; the Beethoven excerpt additionally requires simultaneous keystrokes for octaves and multinote chords, making it the most technically varied of the three excerpts.
Initial trajectories for the excerpts are obtained as described in Section~\ref{sec:train_init_traj}.
Playing accuracy is evaluated using whole-piece F1.

In Stage I, GGN is jointly pretrained on simulated segments from longer right-hand excerpts of the same three pieces: approximately $90$\,s each of Beethoven's Symphony No.~5 and Für Elise, and approximately $120$\,s of Canon. % chktex 13
In Stage II, adaptation is performed independently for each piece: starting from the same pretrained parameters, GGN is fine-tuned only on that piece's execution data.
Stage II is carried out in simulation for the simulation experiments and on the physical hand for the hardware experiments.
In simulation, the Stage II perturbations are more diverse than those used in Stage I and follow a different distribution from the pretraining ranges in Table~\ref{tab:perturbation}, so the prior is not evaluated on its training distribution.
Interaction cost is reported as \emph{simulated environment interaction time} in simulation, excluding training and scheduling, and as \emph{cumulative wall-clock time} on hardware, including these overheads.

% 细节：仿真环境中的阶段二，使用比阶段一更多样且分布不一致的扰动
% 阶段一训练的曲目是约90s的 Beethoven's Symphony No.~5，Für Elise，和约120s的 Canon，这些都来自完整曲目的节选。

\subsection{GGN Design}
\label{sec:exp_design}

We evaluate two design choices: \emph{bidirectional local observation windows} and \emph{correction frequency}.
The evaluation metric is the median F1 improvement after adaptation across $512$ parallel simulated environments during Stage I pretraining,
denoted by $\Delta\mathrm{F1}_{0.5}$.
% 就是在分布内做设计选择

\begin{figure}
    \centering
    \begin{minipage}[b]{0.4\columnwidth}
        \centering
        \includegraphics[width=\linewidth]{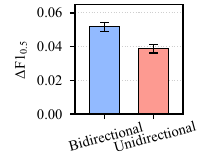}
        \\[-5pt]
        \makebox[\linewidth][c]{\hspace*{0.36in}\subfiglabel{a}}
    \end{minipage}%
    \hfill
    \begin{minipage}[b]{0.58\columnwidth}
        \centering
        \includegraphics[width=\linewidth]{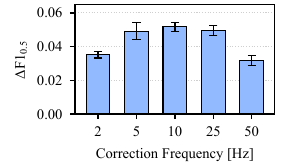}
        \\[-5pt]
        \makebox[\linewidth][c]{\hspace*{0.36in}\subfiglabel{b}}
    \end{minipage}
    \vspace{-10pt}
    \caption{Comparison of GGN design choices.
        (a) Bidirectional versus unidirectional local observation windows.
        (b) Different correction frequencies.
        $\Delta\mathrm{F1}_{0.5}$ is the median F1 improvement across $512$ parallel environments; bars and error bars show the mean and standard deviation over $5$ seeds.}
    \label{fig:exp1_ablation}
    \vspace{-15pt}
\end{figure}

Fig.~\ref{fig:exp1_ablation}(a) compares bidirectional and unidirectional local observation windows. % chktex 36
\textbf{Unidirectional} replaces the observation window in~\eqref{eq:localobs} with $\mathbf{o}_{t_k-f_o:t_k}$,
which retains only information at and before the correction time, with all other training settings unchanged.
The bidirectional window achieves $\Delta\mathrm{F1}_{0.5}$ of $0.052\pm0.003$, compared with $0.039\pm0.003$ for the unidirectional window,
indicating that execution information after the correction time provides useful context for adaptation.

Fig.~\ref{fig:exp1_ablation}(b) compares correction frequencies from $2$ to $50$\,Hz, % chktex 36
corresponding to correction-node intervals $T_c$ from $25$ to $1$ in~\eqref{eq:interp}.
As correction frequency increases, $\Delta\mathrm{F1}_{0.5}$ first rises and then falls, peaking at $10$\,Hz ($T_c{=}5$, the setting used in this work).
Low frequencies may limit the expressiveness of the corrections,
whereas high frequencies enlarge the correction space and make the corrections more prone to jitter.

\begin{figure*}[t]
    \centering
    \includegraphics[width=\textwidth]{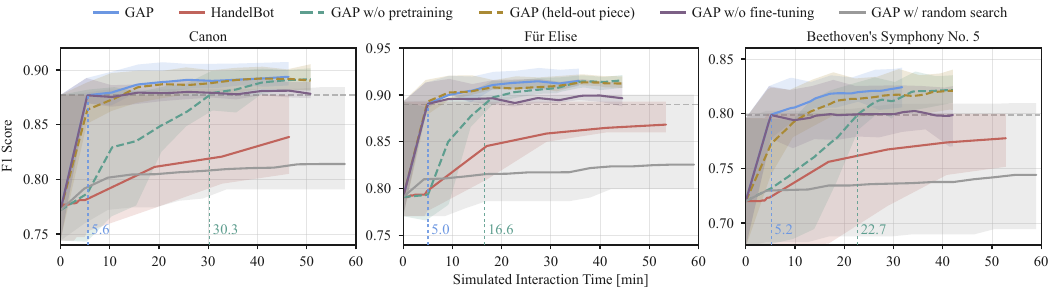}
    \vspace{-20pt}
    \caption{Simulation comparison of GAP. % chktex 13
        Curves show mean whole-piece F1 over $5$ random seeds; shading spans the middle three seeds after removing the best and worst.
        Time denotes simulated environment interaction time, excluding training and scheduling, with $0$ denoting the initial trajectory.
        Dashed lines mark the GAP few-shot F1 and the interaction required by GAP and GAP w/o pretraining to reach it.}
    \label{fig:sim_compare}
    \vspace{-15pt}
\end{figure*}

\subsection{Simulation Comparison Experiments}
\label{sec:exp_sim}

To assess how the generalized adaptation prior contributes to adaptation efficiency,
we compare the full \textbf{GAP} method in simulation with a dexterous piano playing adaptation baseline and four GAP variants.
\textbf{HandelBot}\cite{xie_handelbot_2026} is reimplemented within our simulation framework and serves as the baseline.
\textbf{GAP w/o pretraining} removes Stage I pretraining and starts GGN $F_\theta$ from random initialization
to evaluate the benefit of simulation pretraining.
\textbf{GAP (held-out piece)} excludes all excerpts of the target piece from pretraining, so the target piece is unseen before adaptation; fine-tuning during adaptation is retained.
\textbf{GAP w/o fine-tuning} uses the pretrained GGN but freezes its parameters throughout adaptation; correction generation and iterative execution are retained.
\textbf{GAP w/ random search} retains the same low-frequency correction representation and iterative procedure
but replaces gradient-guided action corrections with random sampling to assess the role of learned adaptation guidance.
Fig.~\ref{fig:sim_compare} summarizes the comparison.
All methods improve playing performance, but GAP achieves larger gains early in adaptation.

\textbf{Adaptation Performance.}
Under the same $20$\,min interaction budget (here and below, F1 at $20$\,min is linearly interpolated between adjacent whole-piece evaluations),
GAP raises the mean F1 score across pieces from the common initial value of $0.761$ to $0.873$, whereas HandelBot reaches $0.807$.
The improvement achieved by GAP is thus approximately $2.4$ times that of HandelBot.
Even without the simulation prior, GAP w/o pretraining reaches $0.844$ and still exceeds HandelBot,
suggesting that part of the gain stems from the adaptation formulation itself.

\textbf{Generalization to Unseen Pieces.}
With the target piece held out from pretraining, GAP (held-out piece) still raises the mean few-shot F1 across pieces from $0.761$ to $0.843$, close to the $0.855$ obtained by GAP, whose pretraining includes the target piece.
Because these gains precede any fine-tuning, they show that experience from other pieces can guide adaptation to a new piece.
With subsequent fine-tuning, its mean F1 across pieces reaches $0.869$ at $20$\,min, close to GAP's $0.873$.
Thus, holding out the target piece from pretraining has little effect on performance at this budget in the evaluated cases.
Together, these results support GAP's ability to learn adaptation guidance shared across pieces.

\textbf{Ablation Analysis.}
The ablations in Fig.~\ref{fig:sim_compare} distinguish the roles of pretraining, continual fine-tuning, and gradient guidance.
Without pretraining, reaching the F1 level attained by GAP after few-shot adaptation requires $23.2$\,min of interaction on average across the pieces, compared with $5.3$\,min for GAP, approximately $4.4$ times as much interaction.
With further interaction, GAP w/o pretraining approaches GAP's F1 on all pieces.
These results indicate that pretraining primarily improves early adaptation efficiency, while adaptation prior knowledge can also be learned from execution data without a pretrained prior.
GAP w/o fine-tuning matches GAP after the first iteration, since both generate corrections using the same pretrained GGN before any Stage II network update.
Its F1 subsequently remains close to the few-shot level, whereas GAP continues to improve.
This contrast supports the role of continual fine-tuning in refining the pretrained prior using new execution feedback when the prior alone provides limited further improvement.
The two stages therefore serve complementary roles: pretraining enables rapid initial adaptation, and fine-tuning supports subsequent gains.
Replacing GGN gradient guidance with random search yields the lowest mean F1 among the evaluated methods under the same $20$\,min interaction budget ($0.785$).
Although random search improves on the initial trajectories, its smaller gains show that the low-frequency correction representation and repeated executions alone do not provide comparable adaptation.
The comparison supports the contribution of GGN gradients to guiding corrections toward improved execution.

\begin{figure}
    \centering
    \vspace{-1pt}
    \includegraphics[width=\columnwidth]{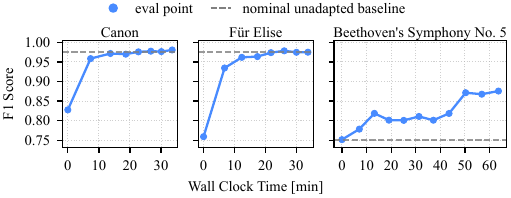}
    \vspace{-22pt}
    \caption{F1 scores at each whole-piece evaluation during real-world adaptation of three representative pieces.
        Evaluations $0$ and $1$ denote the initial and few-shot trajectories, respectively.
        Time includes execution, training, and scheduling. Dashed lines show direct-deployment F1 without added offsets or adaptation.}
    \label{fig:real_f1_iteration}
    \vspace{-15pt}
\end{figure}

\begin{figure*}[t]
    \centering
    \includegraphics[width=\linewidth]{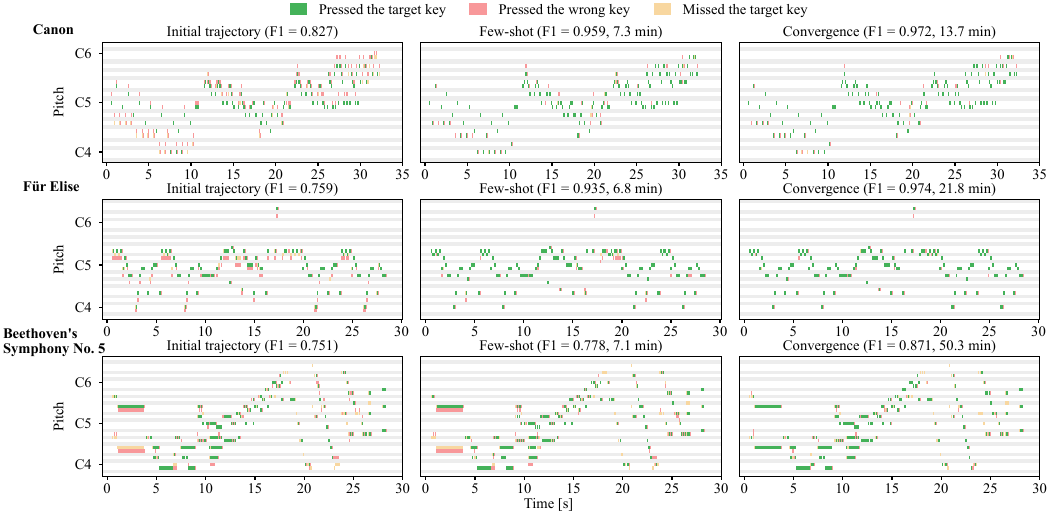}
    \vspace{-25pt}
    \caption{Framewise keystroke results for the three pieces at three stages of real-world adaptation.
        Columns show the initial, few-shot, and converged trajectories, with F1 and cumulative wall-clock time in each panel title.
        Green, red, and orange denote correct, incorrect, and missed keystrokes; gray bands denote black keys.}
    \label{fig:piano_roll_261007}
    \vspace{-10pt}
\end{figure*}

\begin{figure*}[t]
    \centering
    \includegraphics[width=\linewidth]{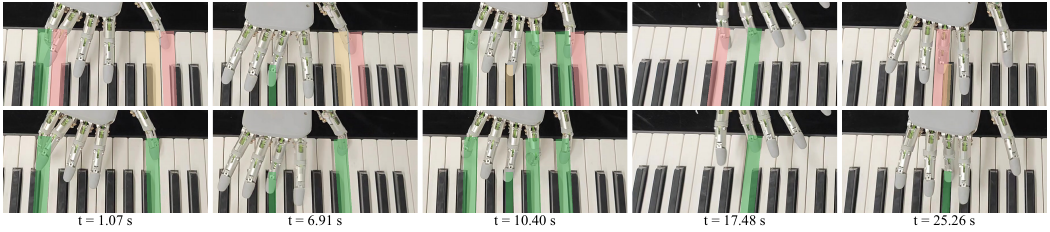}
    \vspace{-25pt}
    \caption{Real-world adaptation by GAP
        at $5$ instants in Beethoven's Symphony No.~5.
        Each column compares the initial (top) and adapted (bottom) keystrokes at one instant.
        Green, red, and orange denote correct, incorrect, and missed keystrokes; examples include off-center, overly light, and mistimed strikes.}
    \label{fig:fig_real_adaptation}
    \vspace{-15pt}
\end{figure*}

\subsection{Real-World Results}
\label{sec:exp_real}

The hardware experiments further examine whether GAP can guide adaptation on the physical system
and whether continued collection of execution data supports subsequent adaptation.
When directly deployed on hardware, the initial trajectories for Canon and Für Elise already achieve F1 scores of approximately $0.975$
(their nominal unadapted baselines), leaving limited room for improvement.
To simulate biases arising from mounting misalignment and systematic errors in real-world settings,
we deliberately apply a global translational offset to the arm (approximately $11$\,mm to the right for Canon and $9$\,mm to the left for Für Elise),
which lies outside the pretraining perturbation ranges (Table~\ref{tab:perturbation}),
reducing their initial F1 scores to $0.827$ and $0.759$, respectively.
Beethoven's Symphony No.~5 achieves an F1 score of $0.751$ upon direct deployment without an artificial offset;
this value is also its nominal unadapted baseline.
Adaptation curves for the three pieces are shown in Fig.~\ref{fig:real_f1_iteration},
and framewise keystrokes at the initial, few-shot, and converged evaluations are shown in Fig.~\ref{fig:piano_roll_261007}.

\textbf{Few-Shot Transfer to Hardware.}
After the first iteration (approximately $7$\,min, before any GGN update with real-world data),
the F1 scores for Canon, Für Elise, and Beethoven's Symphony No.~5 improve by approximately $16\%$, $23\%$, and $3.6\%$, respectively, relative to their initial values.
All three pieces start from the same simulation-pretrained GGN parameters but obtain piece-specific action corrections from their execution observations, supporting transfer of the learned adaptation guidance to hardware.
The larger first-iteration gains for Canon and Für Elise indicate that the generalized adaptation prior can rapidly compensate for the applied global positional offsets.
Subsequently, both pieces exceed their nominal unadapted baselines and stabilize near an F1 score of $0.98$,
indicating that adapted playing accuracy can also surpass that of the initial trajectories without added offsets.

\textbf{Continual Adaptation with Real-World Data.}
Beethoven's Symphony No.~5 exhibits smaller early gains, but continual GGN fine-tuning and gradient-guided adaptation raise its F1 score to $0.871$ by iteration $8$ (approximately $50$\,min). The pretrained prior alone is insufficient to eliminate its playing errors; subsequent adaptation provides further gains. This is consistent with real-world deviations not being fully covered by simulation.

\textbf{Qualitative Analysis of Real-World Adaptation.}
Fig.~\ref{fig:piano_roll_261007} shows framewise playing accuracy for the three pieces.
Incorrect and missed keystrokes in Canon and Für Elise are already substantially reduced during few-shot adaptation, with further reductions afterward.
Beethoven's Symphony No.~5 shows smaller early improvements, with more errors corrected through continual adaptation,
consistent with the trend in whole-piece F1.
Fig.~\ref{fig:fig_real_adaptation} compares keystrokes before and after adaptation at
$5$ instants in the excerpt from Beethoven's Symphony No.~5.
A common error shown is an off-center strike, in which a finger fails to land at the center of a key and may activate an adjacent key.
An off-center finger may even miss the intended black key and become caught on it ($t=25.26$\,s).
Other errors include keystrokes that are too light to activate a key ($t=10.40$\,s)
and reduced accuracy caused by releasing or pressing a key too early or too late ($t=17.48$\,s).
All errors in these examples are corrected after adaptation, indicating that GAP can address playing errors involving \emph{position, key-depression depth, and timing}
and is robust to different types of real-world deviations.

%%%%%%%%%%%%%%%%%%%%%%%%%%%%%%%%%%%%%%%%%%%%%%%%%%%%%%%%%%%%%%%%%%%%%%%%%%%%%%%%
%%%%%%%%%%%%%%%%%%%%%%%%%%%%%%%%%%%%%%%%%%%%%%%%%%%%%%%%%%%%%%%%%%%%%%%%%%%%%%%%
%%%%%%%%%%%%%%%%%%%%%%%%%%%%%%%%%%%%%%%%%%%%%%%%%%%%%%%%%%%%%%%%%%%%%%%%%%%%%%%%
%%%%%%%%%%%%%%%%%%%%%%%%%%%%%%%%%%%%%%%%%%%%%%%%%%%%%%%%%%%%%%%%%%%%%%%%%%%%%%%%
%%%%%%%%%%%%%%%%%%%%%%%%%%%%%%%%%%%%%%%%%%%%%%%%%%%%%%%%%%%%%%%%%%%%%%%%%%%%%%%%
\section{Conclusion}
\label{sec:conclusion}

% 复述问题与方法 + 关键结果。用和 Abstract 一致的主线，但不重复原句，不引入新引用
% 局限性（1–3 条）：明确写出方法/实验的边界
% 未来工作（1–3 条）：由局限自然推出，方向要具体

This letter presented GAP, a sample-efficient adaptation framework that learns a generalized adaptation prior from diverse simulated executions and transfers it to a tendon-driven dexterous hand for piano playing.
In simulation, GAP improved F1 approximately $2.4$ times as much as the existing fast-adaptation method for dexterous piano playing under the same $20$\,min budget.
Without pretraining, reaching GAP's few-shot F1 required approximately $4.4$ times as much interaction.
With the target piece excluded from pretraining, the few-shot F1 remained close to that obtained with the target piece included.
On the physical hand, the same simulation-pretrained prior improved F1 by $3.6\%$--$23\%$ across three representative pieces after approximately $7$\,min each, before any real-world network update, and continual fine-tuning yielded further gains.
These results indicate that how action corrections change performance relative to an observed execution can be learned in simulation and reused across pieces and on hardware.

This work has several limitations.
GAP refines a given initial trajectory, so it benefits from an initial trajectory that already plays most target keys, and its corrections are updated between repeated executions rather than during playing.
The evaluation covers right-hand excerpts of three pieces and measures performance by keystroke accuracy alone.
Future work will extend GAP to two-handed and longer pieces, incorporate musical dynamics and timbre into the reward to improve expressiveness, and explore online correction updates.
Because GAP relies on repeated executions and their feedback rather than on hand-designed correction rules, it can also serve as a sample-efficient adaptation framework for other repetitive real-world manipulation tasks.

\bibliographystyle{IEEEtran}
\bibliography{references}

\end{document}